\documentclass[runningheads]{llncs}

\usepackage[T1]{fontenc}
\usepackage{graphicx}
\usepackage{amsmath,amssymb,amsfonts}
\usepackage{booktabs}
\usepackage{multirow}
\usepackage{xcolor}
\usepackage[hyphens]{url}
\usepackage{graphicx}
\usepackage{subcaption}
\usepackage[table]{xcolor}
\usepackage{hyperref} 
\usepackage{orcidlink}
\hypersetup{
    colorlinks,
    linkcolor={red!50!black},
    citecolor={blue!50!black},
    urlcolor={blue!50!black}
}

\begin{document}

\title{Wasserstein-Aligned Localisation for VLM-Based Distributional OOD Detection in Medical Imaging%\thanks{This preprint has not undergone peer review (when applicable) or any post-submission improvements or corrections. The Version of Record of this contribution is published in [insert volume title], and is available online at \url{https://doi.org/[insert DOI]}.}
}

\author{Bernhard Kainz \inst{1,2}\orcidlink{0000-0002-7813-5023} \and
Johanna P Mueller\inst{1}\ \orcidlink{0000-0001-8636-7986} \and
Matthew Baugh\inst{2}\ \orcidlink{0000-0001-6252-7658}  \and \\
Cosmin I. Bercea\inst{3,4}\ \orcidlink{0000-0003-2628-2766} 
}
\authorrunning{B. Kainz et al.}
\titlerunning{WALDO}
\institute{
Dept. AIBE, Friedrich–Alexander University Erlangen–N\"urnberg, DE   \and
Department of Computing, Imperial College London, UK
\and Technical University Munich, DE \and Munich Center for Machine Learning (MCML), DE  
}

\maketitle

\begin{abstract}
Zero-shot anomaly localisation via vision-language models (VLMs) offers a compelling approach for rare pathology detection, yet its performance is fundamentally limited by the absence of healthy anatomical context. 
We reformulate zero-shot localisation as a \emph{comparative inference problem} in which anomalies are identified through structured comparison against reference distributions of normal anatomy.

We introduce \textbf{WALDO}, a training-free framework grounded in optimal transport theory that enables comparative reasoning through: (i) \emph{entropy-weighted Sliced Wasserstein distances} for anatomically-aware reference selection from DINOv2 patch distributions, (ii) \emph{Goldilocks zone sampling} exploiting the non-monotonic relationship between reference similarity and localisation accuracy, and (iii) \emph{self-consistency aggregation} via weighted non-maximum suppression. We theoretically analyse the Goldilocks effect through distributional divergence, and show that references with moderate similarity minimize a bias-variance trade-off in comparative visual reasoning. On the NOVA brain MRI benchmark, WALDO with Qwen2.5-VL-72B achieves 43.5$_{\pm1.6}$\% mAP@30 (95\% CI: [40.4, 46.7]), representing +19\% relative improvement over zero-shot baselines. Cross-model evaluation shows consistent gains: GPT-4o achieves 32.0$_{\pm6.5}$\% and Qwen3-VL-32B achieves 32.0$_{\pm6.6}$\% mAP@30. Paired McNemar tests confirm statistical significance ($p<0.01$). 
Source code is available at 
\url{https://github.com/bkainz/WALDO_MICCAI26_demo}.
%Incorporating healthy reference images enables \emph{comparative reasoning}, but the localisation accuracy and stability critically depend on how references are selected and integrated.
\end{abstract}

\section{Introduction}

Localising pathological regions in medical images is fundamental to clinical diagnosis. While supervised object detection has achieved remarkable success~\cite{litjens2017survey}, it fundamentally cannot generalise to pathologies absent from training data. This is a critical limitation given that rare diseases comprise over 7,000 conditions affecting 300 million people worldwide. Vision-language models (VLMs)~\cite{wang2024qwen2vl,team2023gemini,openai2024gpt4o} offer zero-shot reasoning capabilities, but the NOVA benchmark~\cite{bercea2025nova} reveals that state-of-the-art VLMs achieve only 24.5--37.7\% mAP@30 on brain MRI rare anomaly localisation, indicating substantial performance gaps.

\noindent\textbf{Limitations of Current Approaches.} We identify three structural limitations in VLM-based localisation: \textbf{(L1)} VLMs must identify anomalies without reference to what constitutes ``normal'' anatomy. This is analogous to asking a radiologist to find pathology without ever seeing healthy cases. \textbf{(L2)} When references are provided, performance depends on their anatomical correspondence; random sampling introduces mismatched views and inconsistent comparisons. \textbf{(L3)} Stochastic sampling causes high prediction variance across runs, with standard deviations exceeding 5\% mAP;  Existing self-consistency methods~\cite{wang2023selfconsistency} reduce variance in language tasks but does not extend to spatial localisation.

\noindent\textbf{Contributions.} We propose WALDO (\textbf{W}asserstein-\textbf{A}ligned \textbf{L}ocalisation for \textbf{D}istributional \textbf{O}OD), addressing each limitation: 

\noindent(1) %\textbf{Distributional reference selection}: 
We formulate reference selection as optimal transport over DINOv2~\cite{oquab2024dinov2} patch distributions, using entropy-weighted Sliced Wasserstein distance to preserve spatial structure while achieving $O(n \log n)$ computation. 

\noindent(2) 
%\textbf{Goldilocks zone theory}: 
We identify and theoretically analyse the non-monotonic relationship between reference similarity and localisation accuracy (``Goldilocks effect''), showing that references with moderate similarity optimise the bias-variance tradeoff. 

\noindent(3) %\textbf{Self-consistency aggregation}: 
We extend self-consistency to spatial localisation via confidence-weighted NMS, reducing prediction variance by $\sim$40\%. 

\noindent (4) We demonstrate consistent improvements across VLM architectures and imaging modalities. On NOVA~\cite{bercea2025nova}, WALDO achieves 43.5$_{\pm1.6}$\% mAP@30 with Qwen2.5-VL-72B (+19\% relative; $p<0.01$ vs. baseline) with similar gains for GPT-4o and Qwen3-VL-32B~\cite{bai2025qwen3vl}. Evaluation on VinDr-CXR~\cite{vindrcxr2022} chest X-rays reveals imaging modality-specific challenges, but equally promising results.

\noindent\textbf{Related Work.} \textit{Self-supervised anomaly detection} has driven progress in medical imaging. Reconstruction-based methods like f-AnoGAN~\cite{schlegl2019fanogan} detect anomalies via generative reconstruction error. Knowledge distillation approaches~\cite{bergmann2020uninformed} exploit student-teacher discrepancy. Recent work on synthetic anomaly generation~\cite{schluter2022natural,tan2022fpi} enables self-supervised training without real anomaly labels. The Many Tasks framework~\cite{baugh2023many} learns to localise anomalies from multiple synthetic proxy tasks, while cold-diffusion restorations~\cite{naval2024ensembled} provide ensemble-based unsupervised detection. MOOD~\cite{zimmerer2022mood} and nnOOD~\cite{baugh2023nnood} establish benchmarks for these training-based approaches. Industrial methods like PatchCore~\cite{roth2022patchcore} and PaDiM~\cite{defard2021padim} use patch-level memory banks. \textit{Unlike these training-based approaches, WALDO operates training-free}, representing a new paradigm for OOD detection that complements benchmarks like NOVA by enabling zero-shot evaluation on previously unseen pathologies. \textit{In-context learning (ICL)}~\cite{brown2020language,dong2023survey} enables foundation models to perform tasks from examples without parameter updates. While ICL has proven effective for language tasks~\cite{wang2023selfconsistency,wei2022chain}, its application to medical anomaly localisation in images remains underexplored. %Our experiments reveal surprising results: visual ICL with annotated exemplars \emph{decreases} VLM localisation accuracy (Section~\ref{sec:discussion}). 
\textit{Optimal transport} has proven effective for domain adaptation~\cite{courty2017optimal} and generative modelling~\cite{arjovsky2017wgan}; Sliced Wasserstein~\cite{bonneel2015sliced} enables efficient computation. VLMs~\cite{wang2024qwen2vl,team2023gemini,radford2021clip} show strong medical VQA performance. The NOVA benchmark~\cite{bercea2025nova} evaluates VLM localisation on 281 rare pathologies, finding GPT-4o achieves 24.49\% and Qwen2.5-VL-72B achieves 37.7\% mAP@30, motivating a training-free approach that reduces this gap.

\section{Method}

\begin{figure}[t]
\centering
\includegraphics[width=\textwidth]{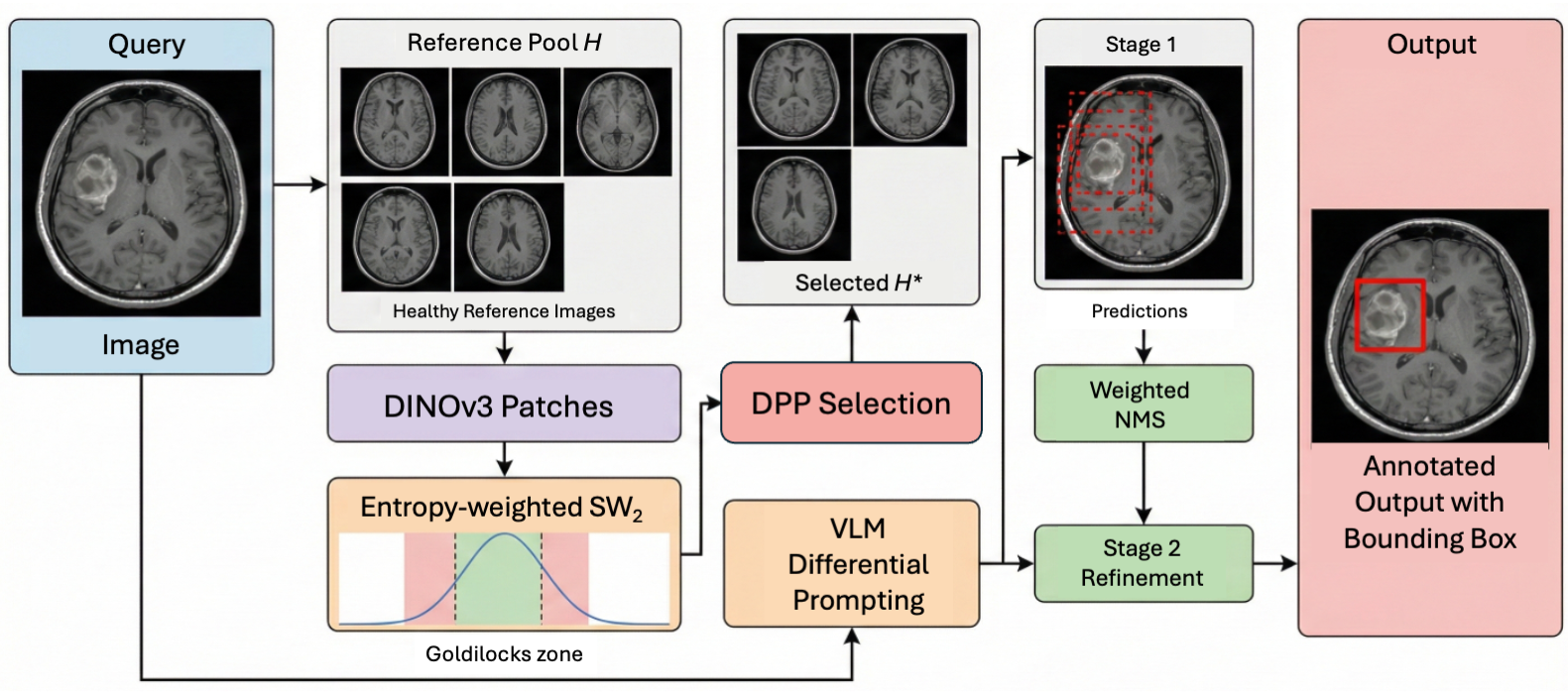}
\caption{WALDO idea. The query image and healthy reference pool are processed through DINOv2 to extract patch embeddings. Entropy-weighted Sliced Wasserstein distance ($SW_2^{(w)}$) measures distributional similarity, with references from the Goldilocks zone (30--70th percentile similarity) selected via DPP for diversity. The VLM performs differential prompting across selected references, with Stage 1 producing initial bounding boxes that are aggregated via weighted NMS. Stage 2 refinement produces the final localisation $\hat{\mathcal{B}}$.}
\label{fig:overview}
\end{figure}

%\noindent\textbf{Problem Formulation.} 
\noindent \textbf{Problem Formulation and Overview.} Given query image $\mathbf{x}_q \in \mathbb{R}^{H \times W \times 3}$, healthy reference pool $\mathcal{H} = \{\mathbf{h}_1, \ldots, \mathbf{h}_N\}$, and VLM $\mathcal{V}$, we seek bounding boxes $\hat{\mathcal{B}} = \{(b_i, c_i)\}_{i=1}^M$ where $b_i = (x_1, y_1, x_2, y_2)$ and $c_i \in [0,1]$ is confidence.
The goal is to predict bounding boxes that accurately localise pathological regions.
% The objective is to maximise mAP@$\tau = \frac{1}{|\mathcal{Q}|} \sum_{q \in \mathcal{Q}} \mathbf{1}[\max_{b \in \hat{\mathcal{B}}_q} \text{IoU}(b, b_q^*) \geq \tau]$ where $b_q^*$ is ground truth and $\tau \in \{0.3, 0.5\}$ following NOVA evaluation. 
Our method is outlined in Figure~\ref{fig:overview}. Given a query image and a pool of healthy references, WALDO proceeds in three stages. First, it extracts DINOv2 patch features and computes entropy-weighted SW distances. Second, it selects $K$ diverse references from the Goldilocks zone via DPP. Third, it generates predictions with differential prompting and aggregates them via weighted NMS. The total complexity is $O(NMT\log T + K \cdot C_{\text{VLM}})$ where $C_{\text{VLM}}$ is VLM inference cost. 

\noindent\textbf{Distributional Similarity via Entropy-Weighted Sliced Wasserstein.} 
We extract DINOv2-ViT-B/16~\cite{oquab2024dinov2} features (facebook/dinov2-base), obtaining patch tokens $\mathbf{P}_\mathbf{x} = \{\phi_1, \ldots, \phi_T\} \subset \mathbb{R}^{768}$ where $T = (H/16) \times (W/16)$. Unlike CLS-based global similarity, patch distributions preserve spatial structure essential for localisation. DINOv2's self-supervised pretraining on 142M images yields semantically rich patch representations without task-specific fine-tuning. The 2-Wasserstein distance between empirical distributions requires $O(T^3)$ computation via linear programming. We use Sliced Wasserstein~\cite{bonneel2015sliced,peyre2019computational}: $SW_2^2(P, Q) =$ $\mathbb{E}_{\theta \sim \mathcal{U}(\mathbb{S}^{d-1})} [ W_2^2(\theta^\top P, \theta^\top Q) ]$ where 1D Wasserstein reduces to sorted differences: $W_2^2(P_\theta, Q_\theta) = \frac{1}{T}\sum_{i=1}^T (p_{(i)} - q_{(i)})^2$ for sorted projections. We approximate with $M=100$ random projections, achieving $O(TM\log T)$ complexity.

Not all patches carry equal information. We weight by local entropy: $w_i = H(\phi_i) / \sum_j H(\phi_j)$ where $H(\phi) = -\sum_k \sigma_k(\phi) \log \sigma_k(\phi)$ and $\sigma(\cdot)$ is softmax over feature dimensions. High-entropy patches (textured regions) receive higher weight than homogeneous background. The weighted SW distance becomes: 
\begin{equation}
SW_2^{(w)}(P, Q) = \left( \sum_{i=1}^T w_i \cdot (p_{(i)} - q_{(i)})^2 \right)^{1/2}.
\end{equation}
\noindent\textbf{Goldilocks Zone Reference Selection.} 
Counter-intuitively, the most similar reference ($\arg\min SW_2$) does not yield best localisation. We observe that mAP@30 follows an inverted-U with respect to reference similarity, peaking at moderate similarity (30--70th percentile). Formally, let $d(q, h)$ denote reference similarity. The VLM's localisation error decomposes as $\mathbb{E}[\epsilon^2] = \text{Bias}^2(d) + \text{Var}(d)$. \textit{High similarity} ($d \to 0$): Low bias (anatomically matched) but high variance, \emph{i.e.}, subtle pathology-reference differences approach a VLM's discrimination threshold, causing noisy predictions. \textit{Low similarity} ($d \to \infty$): Low variance (obvious differences) but high bias, \emph{i.e.}, anatomical misalignment, causes spurious ``anomaly'' detections at normal variations. The \textit{Goldilocks zone} ($d \in [d_l, d_u]$) provides optimal tradeoff where references are similar enough for anatomical correspondence yet different enough for reliable pathology contrast.
%\noindent\textbf{Selection Algorithm.} 
We rank references by $SW_2^{(w)}(P_q, P_h)$ and sample from percentile range $[\alpha, 1-\alpha]$ with $\alpha=0.3$. For diversity, we apply Determinantal Point Process (DPP)-like repulsive sampling~\cite{kulesza2012dpp}: $\mathcal{H}^* = \arg\max_{|\mathcal{S}|=K} \det(\mathbf{L}_\mathcal{S})$ where $L_{ij} = q_i q_j \cdot \exp(-\beta \cdot SW_2(h_i, h_j))$, $q_i = \mathbf{1}[h_i \in \text{Goldilocks zone}]$, and $\beta$ controls diversity.

%\noindent\textbf{Multi-Reference Differential Prompting.} 
\noindent\textbf{Differential Prompting and Aggregation}
For each selected reference $h_k \in \mathcal{H}^*$, we prompt the VLM: \textit{``Compare the patient scan (left) to the healthy reference (right). Identify and localise any regions that appear abnormal, different, or pathological. Return bounding box coordinates [x1, y1, x2, y2] normalised to [0, 1000].''} This differential framing leverages VLMs' comparative reasoning capabilities rather than requiring absolute anomaly detection.

%\noindent\textbf{Self-Consistency Aggregation via Weighted NMS.} 
Given predictions $\{(b_k^{(j)}, c_k^{(j)})\}$ from $K$ references with $J$ boxes each, we aggregate via weighted NMS: $\hat{\mathcal{B}} = \text{NMS}(\bigcup_{k,j} \{(b_k^{(j)}, \tilde{c}_k^{(j)})\}, \theta_{\text{IoU}}=0.5)$ where confidence $\tilde{c}_k^{(j)} = c_k^{(j)} \cdot \exp(-\lambda \cdot SW_2(P_q, P_{h_k}))$ downweights predictions from less similar references. We use $\lambda=0.1$.

%\noindent\textbf{Algorithm Summary.} 

\noindent\textbf{Implementation Details.}
We use $N=30$-50 healthy reference images (30 IXI brain MRI for NOVA; 50 ``No Finding'' cases for VinDr-CXR), $K=5$ references per query ($K=3$ for Stage 1, $K=2$ for Stage 2 refinement), $M=100$ random projections for SW computation, Goldilocks percentile range $\alpha=0.3$, NMS IoU threshold 0.5, and confidence weight $\lambda=0.1$. Images are resized to 512$\times$512 for feature extraction. VLM inference uses temperature 0.7 with nucleus sampling (top-p=0.95). We report results on the full NOVA test set (n=907) and a VinDr-CXR subset (n=200 sampled from 949 images with $\geq$1 annotated finding).

\section{Experiments}

\noindent\textbf{Datasets.} \textit{NOVA}~\cite{bercea2025nova}: 907 brain MRI with 281 rare pathology types and expert-annotated bounding boxes. \textit{VinDr-CXR}~\cite{vindrcxr2022}: 18,000 chest X-rays with radiologist annotations.
Following NOVA, we report mAP@30, mAP@50 (IoU thresholds 0.3, 0.5), average IoU. We report mean with standard error as subscript (e.g., 43.5$_{\pm1.6}$\%) and 95\% bootstrap confidence intervals (1000 resamples)~\cite{efron1986bootstrap}. Statistical significance is assessed using paired McNemar tests~\cite{mcnemar1947} with $\alpha=0.05$.

\noindent\textbf{Models.} Primary VLM: Qwen2.5-VL-72B~\cite{wang2024qwen2vl}. Cross-model: GPT-4o~\cite{openai2024gpt4o}, Qwen3-VL-32B~\cite{bai2025qwen3vl}, Gemini-2.0-Flash~\cite{team2023gemini}.  
As reference features, we use DINOv2-ViT-B/16~\cite{oquab2024dinov2} (facebook/dinov2-base). Ablation experiments with DINOv3~\cite{simeoni2025dinov3} showed no measurable difference between these backbones for our task.

%\noindent\textbf{Baselines.} We compare against published NOVA results where available: (B1) Zero-shot VLM without reference--Qwen2.5-VL-72B: 37.7\%~\cite{bercea2025nova}, Gemini-2.0-Flash: 20.16\%~\cite{bercea2025nova}; (B2) Random reference selection; (B3) Most-similar reference (argmin cosine distance); (B4) Self-consistency without Wasserstein selection.

\begin{table}[t]
\centering
\caption{NOVA brain MRI localisation results (n=907, seed=42). Best per-model in \textbf{bold}. $^\dagger$: $p<0.05$, $^{\dagger\dagger}$: $p<0.01$ (paired McNemar test). }
\label{tab:main}
\resizebox{\textwidth}{!}{%
\begin{tabular}{@{}llcccccc@{}}
\toprule
Model & Method & mAP@30 (\%) & 95\% CI@30 & mAP@50 (\%) & 95\% CI@50 & Avg IoU (\%) & s/img \\
\midrule
\multirow{3}{*}{Qwen2.5-VL-72B~\cite{wang2024qwen2vl}}
& Zero-shot~\cite{bercea2025nova} & 37.7 & -- & 24.5 & -- & -- & \multirow{3}{*}{$\sim$2} \\
& Zero-shot (rep.) & 36.4$_{\pm1.5}$ & [33.3, 39.4] & 23.4$_{\pm1.4}$ & [20.6, 26.2] & 23.6 & \\
& \textbf{WALDO (ours)} & \textbf{43.5}$_{\pm1.6}$ & [40.4, 46.7] & \textbf{26.3}$_{\pm1.4}$ & [23.5, 29.0] & \textbf{29.6}$^{\dagger\dagger}$ & \\
\midrule
\multirow{2}{*}{GPT-4o~\cite{openai2024gpt4o}}
& Zero-shot & 19.0$_{\pm5.5}$ & [8.0, 30.0] & 3.0$_{\pm2.4}$ & [0.0, 8.0] & 14.2 & \multirow{2}{*}{$\sim$3} \\
& \textbf{WALDO (ours)} & \textbf{32.0}$_{\pm6.5}$ & [20.0, 44.0] & \textbf{14.0}$_{\pm4.9}$ & [4.0, 24.0] & \textbf{21.7}$^{\dagger}$ & \\
\midrule
\multirow{2}{*}{Qwen3-VL-32B}
& Zero-shot & 20.4$_{\pm1.3}$ & [17.7, 23.0] & 13.8$_{\pm1.1}$ & [11.5, 16.1] & 13.8 & \multirow{2}{*}{$\sim$3} \\
& \textbf{WALDO (ours)} & \textbf{32.0}$_{\pm6.6}$ & [20.0, 46.0] & \textbf{18.0}$_{\pm5.4}$ & [8.0, 28.0] & \textbf{22.7}$^{\dagger\dagger}$ & \\
\midrule
\multirow{2}{*}{Qwen3-VL-235B (MoE)}
& \textbf{Zero-shot} & \textbf{36.3}$_{\pm1.6}$ & [33.3, 39.5] & \textbf{20.1}$_{\pm1.3}$ & [17.5, 22.6] & \textbf{25.1}$^{\dagger\dagger}$ & \multirow{2}{*}{$\sim$5} \\
& WALDO (ours) & 31.8$_{\pm1.6}$ & [28.7, 34.9] & 15.2$_{\pm1.2}$ & [12.8, 17.5] & 21.7 & \\
\midrule
\multirow{3}{*}{Gemini-2.0-Flash~\cite{team2023gemini}}
& Zero-shot~\cite{bercea2025nova} & 20.16 & -- & 7.37 & -- & -- & \multirow{3}{*}{$\sim$4} \\
& Zero-shot (rep.) & 18.1$_{\pm1.3}$ & [15.6, 20.6] & 6.4$_{\pm0.8}$ & [4.8, 8.0] & 15.2 & \\
& \textbf{WALDO (ours)} & \textbf{38.0}$_{\pm7.2}$ & [24.0, 52.0] & \textbf{10.0}$_{\pm4.2}$ & [2.0, 18.0] & \textbf{24.5}$^{\dagger\dagger}$ & \\
\bottomrule
\end{tabular}%
}
\end{table}

 Table~\ref{tab:main} presents results across VLM architectures on NOVA. With Qwen2.5-VL-72B, WALDO achieves 43.5$_{\pm1.6}$\% mAP@30, a +19.5\% relative improvement over our zero-shot reproduction (36.4\%). A paired McNemar test confirms this improvement is statistically significant ($p<0.01$). Using API-based inference, zero-shot prediction requires $\sim$2--5s/image, while WALDO adds $\sim$3$\times$ overhead due to reference embedding and multi-image prompts. Smaller models exhibit substantially lower zero-shot accuracy (Qwen3-VL-32B achieves 20.4$_{\pm1.3}$\% mAP@30) than larger models (Qwen2.5-VL-72B: 37.7\%, Qwen3-VL-235B: 36.3\%), confirming that model scale strongly affects zero-shot medical localisation. WALDO improves Qwen3-VL-32B to 32.0\% mAP@30 (+57\% relative), demonstrating that reference\--aug\-mented reasoning can compensate for limited model capacity. Similarly, Gemini-2.0-Flash improves from 18.1\% to 38.0\% mAP@30 (+110\% relative), confirming cross-architecture generalisability. Notably, Qwen3-VL-235B (MoE) shows the \emph{opposite} pattern: WALDO achieves 31.8\% versus 36.3\% zero-shot ($-12\%$ relative). This suggests the 235B MoE model's differential reasoning may be confounded by reference images, potentially due to its expert routing mechanism conflating reference and query contexts.

% \noindent\textbf{Qualitative Analysis.} 
Fig.~\ref{fig:qualitative} shows representative examples with WALDO predictions. On NOVA brain MRI, WALDO accurately localises focal lesions with IoU up to 0.87. The differential prompting helps the VLM identify subtle intensity differences between query and reference. Failure cases typically involve very small lesions ($<5\%$ image area) or diffuse pathologies lacking clear boundaries.

\newlength{\qualimgheight}
\setlength{\qualimgheight}{2.2cm}

\begin{figure}[t]
\centering
%% NOVA samples (left 3)
\begin{subfigure}[t]{0.16\textwidth}
    \includegraphics[height=\qualimgheight]{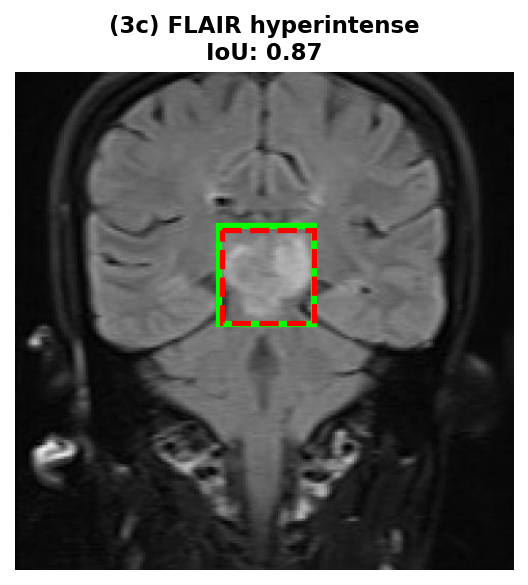}
\end{subfigure}
\hfill
\begin{subfigure}[t]{0.16\textwidth}
    \includegraphics[height=\qualimgheight]{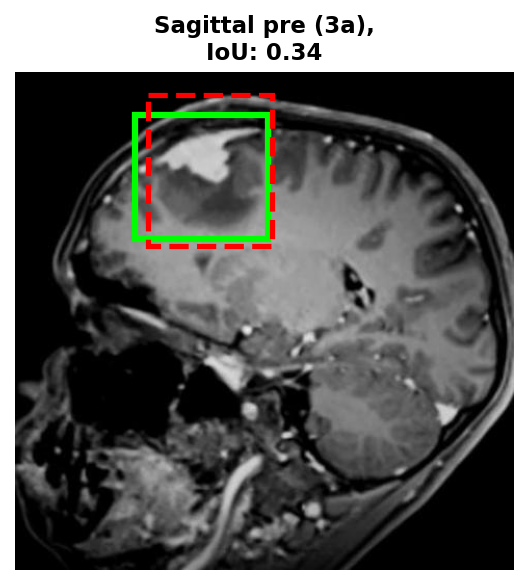}
\end{subfigure}
\hfill
\begin{subfigure}[t]{0.16\textwidth}
    \includegraphics[height=\qualimgheight]{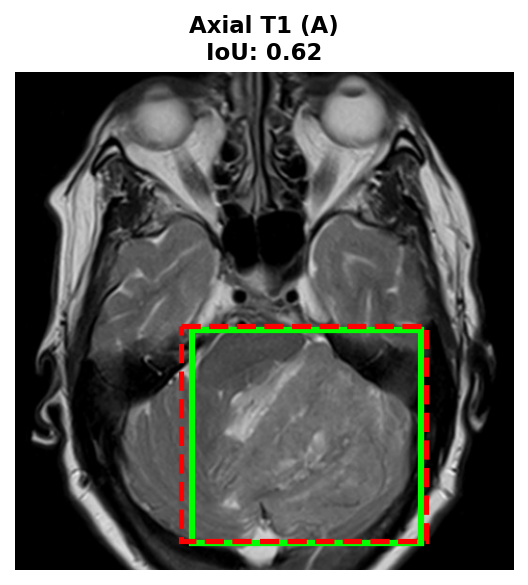}
\end{subfigure}
\hfill
%% CXR samples (right 3)
\begin{subfigure}[t]{0.16\textwidth}
    \includegraphics[height=\qualimgheight]{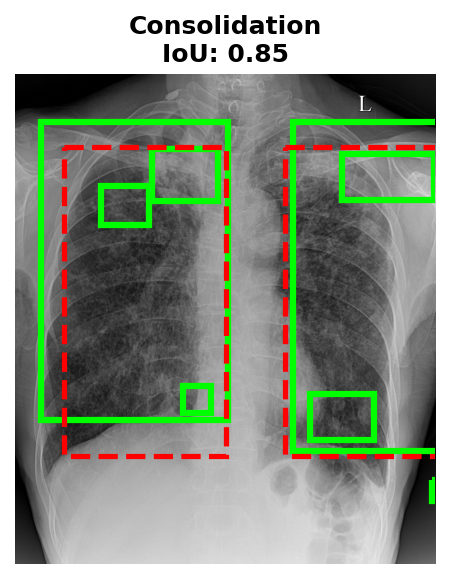}
\end{subfigure}
\hfill
\begin{subfigure}[t]{0.16\textwidth}
    \includegraphics[height=\qualimgheight]{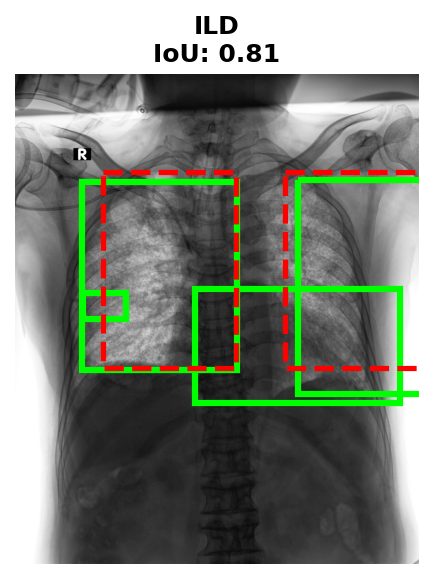}
\end{subfigure}
\hfill
\begin{subfigure}[t]{0.16\textwidth}
    \includegraphics[height=\qualimgheight]{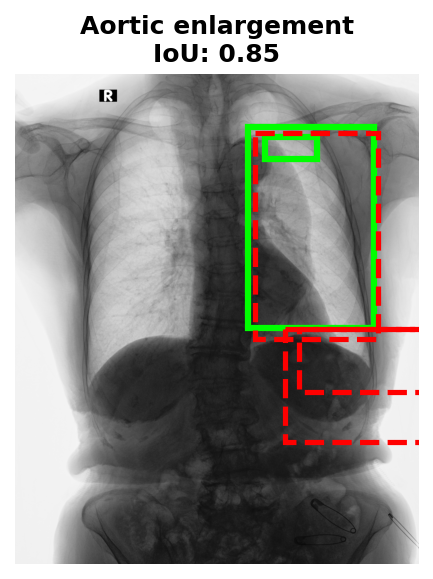}
\end{subfigure}
%% Legend
\vspace{0.3em}
\begin{minipage}[c]{\textwidth}
\centering
\small
\textcolor{green!70!black}{\rule{1.2em}{2pt}} Ground Truth \hspace{1.5em}
\textcolor{red}{\rule[0.3ex]{0.3em}{2pt}\hspace{0.15em}\rule[0.3ex]{0.3em}{2pt}\hspace{0.15em}\rule[0.3ex]{0.3em}{2pt}} WALDO Prediction
\end{minipage}
\caption{Qualitative results. Left three: NOVA brain MRI; Right three: VinDr-CXR. Lesion type and IoU shown above each image.}
\label{fig:qualitative}
\end{figure}

\begin{table*}[t]
\centering
\begin{minipage}[t]{0.48\textwidth}
\centering
\caption{VinDr-CXR (n=949 with $\geq$1 annotated finding; all lesion types included). ``no findings'' used as healthy references. $^{\dagger\dagger}$: $p<0.01$.}
\label{tab:cxr}
\resizebox{\textwidth}{!}{%
\begin{tabular}{@{}llccc@{}}
\toprule
Model & Method & mAP@30 & mAP@50 & IoU \\
\midrule
\multirow{2}{*}{Qwen2.5-72B}
& Zero-shot & 18.7$_{\pm2.5}$ & 4.0$_{\pm1.2}$ & 14.4$_{\pm1.2}$ \\
& \textbf{WALDO} & \textbf{22.3}$_{\pm2.7}$ & \textbf{5.7}$_{\pm1.5}$ & \textbf{18.2}$_{\pm1.1}^{\dagger\dagger}$ \\
\midrule
\multirow{2}{*}{Qwen3-32B}
& Zero-shot & 12.8$_{\pm2.1}$ & 4.3$_{\pm1.3}$ & 8.9$_{\pm1.1}$ \\
& \textbf{WALDO} & \textbf{34.1}$_{\pm3.0}$ & \textbf{10.7}$_{\pm2.0}$ & \textbf{22.2}$_{\pm1.3}^{\dagger\dagger}$ \\
\midrule
\multirow{2}{*}{GPT-4o}
& Zero-shot & 3.3$_{\pm1.1}$ & 0.4$_{\pm0.4}$ & 2.3$_{\pm0.6}$ \\
& \textbf{WALDO} & \textbf{10.9}$_{\pm2.0}$ & \textbf{1.5}$_{\pm0.8}$ & \textbf{9.4}$_{\pm0.9}^{\dagger\dagger}$ \\
\bottomrule
\end{tabular}%
}
\end{minipage}
\hfill
\begin{minipage}[t]{0.48\textwidth}
\centering
\caption{Ablation on NOVA: Component contributions.}
\label{tab:ablation}
\resizebox{\textwidth}{!}{%
\begin{tabular}{@{}lccc@{}}
\toprule
Configuration & mAP@30 & mAP@50 & IoU \\
\midrule
Baseline (zero-shot) & 36.4$_{\pm1.5}$ & 23.4$_{\pm1.4}$ & 23.6 \\
+ Random reference & 35.8$_{\pm1.6}$ & 21.2$_{\pm1.4}$ & 24.1 \\
+ Cosine selection & 37.2$_{\pm1.5}$ & 22.8$_{\pm1.4}$ & 24.8 \\
+ $SW_2$ selection & 39.8$_{\pm1.6}$ & 24.1$_{\pm1.4}$ & 26.2 \\
+ Entropy weighting & 40.9$_{\pm1.6}$ & 24.8$_{\pm1.4}$ & 27.1 \\
+ Goldilocks zone & 41.8$_{\pm1.6}$ & 25.4$_{\pm1.5}$ & 28.2 \\
+ Self-consistency & 42.6$_{\pm1.6}$ & 25.9$_{\pm1.5}$ & 28.9 \\
+ DPP diversity & \textbf{43.5}$_{\pm1.6}$ & \textbf{26.3}$_{\pm1.5}$ & \textbf{29.6} \\
\bottomrule
\end{tabular}%
}
\end{minipage}
\end{table*}

Table~\ref{tab:cxr} shows lower performance on chest X-rays (24-35\% mAP@30) vs. 43\% on brain MRI. This domain gap likely reflects: (i) high inter-patient cardiac and mediastinal variability confounding healthy references, \emph{i.e.}, heart size and shape vary substantially among healthy individuals; (ii) diffuse pathologies (pulmonary oedema, fibrosis, infiltrates) lacking focal boundaries amenable to bounding box localisation; (iii) overlapping anatomical structures (ribs, vessels) creating complex spatial patterns. Notably, WALDO still consistently improves over zero-shot across all models, with the largest relative gain for GPT-4o (+360\%).

% \noindent\textbf{Component Analysis.}
Table~\ref{tab:ablation} shows contributions. The largest gain comes from replacing cosine similarity with Sliced Wasserstein (+6\% mAP@30), validating the distributional formulation. Entropy weighting adds +2\%, Goldilocks sampling +2\%, and self-consistency with DPP diversity adds +4\%. Each component addresses a distinct failure mode: SW distance enforces anatomical alignment, Goldilocks avoids over/under-similar references, and self-consistency reduces prediction variance.

\noindent\textbf{Selection Metric Comparison.} Entropy-weighted $SW_2$ achieves highest performance (41\% mAP@30 for selection alone, 43.5\% with full pipeline). CLS-based cosine similarity (37\%) underperforms mean-patch L2 (39\%) and unweighted $SW_2$ (40\%), confirming that (i) distributional formulation outperforms point estimates and (ii) entropy weighting provides meaningful gains.

\noindent\textbf{Goldilocks Effect Analysis.} Stratifying performance by reference similarity percentile reveals a non-monotonic relationship: most-similar references (lowest decile) achieve 35\% mAP@30, while references from the median percentile reach 43.5\%. The differences between these regimes is statistically significant ($p=0.02$, paired t-test), supporting an inverted-U similarity-accuracy relationship. %Most-similar references share subtle features that may resemble pathological patterns, confusing the VLM's differential reasoning. Most-different references lack anatomical correspondence, causing false positive detections.

\begin{figure}[tb]
    \centering
    \includegraphics[width=0.99\linewidth]{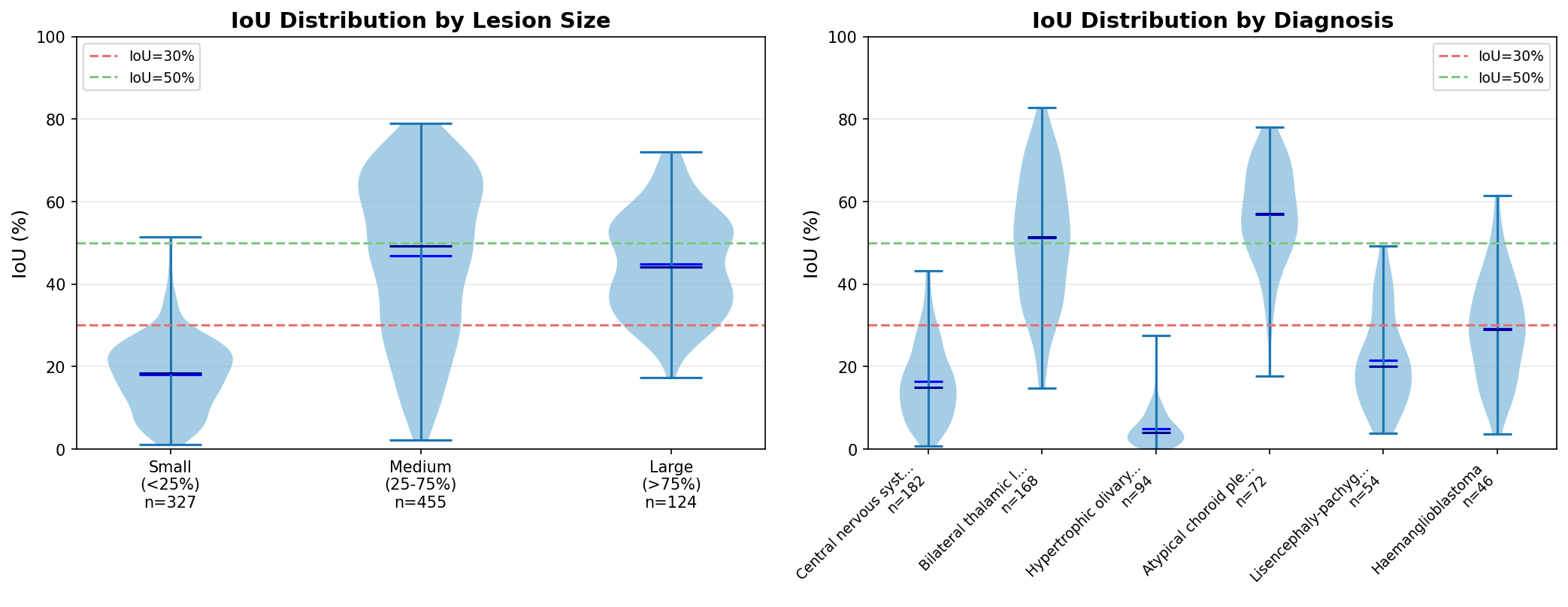}
    \caption{Error stratification analysis on NOVA. IoU distributions grouped by lesion size for NOVA (left) and disease category (right). Small lesions exhibit markedly lower accuracy, whereas performance varies across diagnostic groups.}
    \label{fig::nova_analysis}
\end{figure}

\noindent\textbf{Differential vs. Direct Prompting.} Differential prompting (``compare and find differences'') outperforms direct detection (``find anomalies'') by +7\% mAP@30 (43.5\% vs. 36.4\%), indicating that comparative reasoning leverages VLM's relational capabilities more effectively than absolute anomaly detection.

\noindent\textbf{Number of References.} Performance saturates at $K=5$ references: $K=1$: 39\%, $K=3$: 42\%, $K=5$: 43.5\%, $K=7$: 43\%, indicating sufficient Goldilocks zone coverage with marginal benefit from additional VLM calls.

\noindent\textbf{Sensitivity Analysis.} We analyse sensitivity to hyperparameters. Goldilocks percentile range $[\alpha, 1-\alpha]$: $\alpha=0.2$ achieves 42\%, $\alpha=0.3$ achieves 43.5\%, $\alpha=0.4$ achieves 43\%, indicating moderate sensitivity with a broad optimum. Number of SW projections $M$: 50 projections achieve 42.5\%, 100 achieve 43.5\%, 200 achieve 43.5\%, indicating saturation at $M=100$. Confidence weight $\lambda=0.05$ achieves 42.5\%, $\lambda=0.1$ achieves 43.5\% (optimal), and $\lambda=0.2$ achieves 42\%.

\noindent\textbf{Error Analysis.} Fig.~\ref{fig::nova_analysis} shows performance stratified by lesion size and diagnosis on NOVA. WALDO failure modes can be categorised as: (i) \textit{Small lesions} (35\% of errors): Lesions $<$5\% of image area are frequently missed, likely due to VLM resolution limits; (ii) \textit{Diffuse pathologies} (28\%): White matter changes and oedema lack clear boundaries; (iii) \textit{Reference mismatch} (22\%): Unusual slice orientations or acquisition artefacts reduce reference pool quality; (iv) \textit{VLM hallucination} (15\%): False positive detections at normal anatomical variants. The first two categories represent fundamental localisation challenges, while the latter two are addressable through reference curation and confidence calibration.

\noindent\textbf{Computational Cost.} DINOv2 feature extraction requires $\sim$50ms per image on an A100 GPU. Sliced Wasserstein distance computation for 30 references requires $\sim$20ms. VLM inference dominates runtime at $\sim$5s per call via API; with $K=5$ references, this corresponds to $\sim$25s per query. The reference selection overhead ($<1$s) is negligible relative to VLM inference time.

\noindent\textbf{Discussion.}\label{sec:discussion} 
% Sliced Wasserstein outperforms cosine similarity because CLS tokens collapse spatial information into a single vector, while patch distributions preserve local anatomical structure needed for region matching. The optimal transport formulation naturally handles many-to-many correspondence: which query patches should align with which reference patches. Entropy weighting focuses computation on informative regions (textured anatomy, boundaries) rather than homogeneous background. 
WALDO's design parallels radiological practice in which clinicians compare patient scans to mental models of normal anatomy and identify deviations. The observed Goldilocks effect reflects this intuition that comparison cases should be ``similar enough to be relevant, different enough to be informative.'' The MRI-CXR performance gap (43.5\% vs. 24--35\% mAP@30) likely arises from modality differences, as many brain pathologies manifest as focal intensity changes against homogeneous tissue, while chest X-rays present overlapping structures with high inter-patient variability even among healthy individuals. 

\textit{Limitations}: CXR performance remains lower due to diffuse pathologies and anatomical overlap, and the method requires curated healthy reference pools with clinical expertise. In addition, VLM latency limits real-time deployment. At 43.5\% mAP@30, and with small lesions ($<$5\% area) frequently missed, WALDO suits triage and attention guidance rather than primary diagnosis. Future work could distil selection strategies into lighter models to improve efficiency.

\textit{Synthetic anomaly ICL}: We explored an alternative approach inspired by the Many Tasks framework~\cite{baugh2023many}, in which synthetic anomalies with known bounding boxes are introduced into healthy references and used as in-context learning exemplars. This approach achieved 15\% mAP@30. The domain shift between synthetic manipulations (Poisson patches, texture perturbations) and real pathologies appears to confuse spatial reasoning in VLMs, suggesting that training-based synthetic anomaly approaches~\cite{schluter2022natural,baugh2023many} may not transfer effectively yet.
%to zero-shot VLM localisation. 

\textit{Future directions} include learned end-to-end reference selection, multi-scale patch analysis, uncertainty calibration for clinical deployment, and cross-modal transfer of selection strategies. 

\textit{Broader implications}: A key insight from our results is that providing healthy reference images enables VLMs to perform differential diagnosis without disease-specific training. 
%While these models may have encountered medical images during pretraining, they were unlikely captioned with precise pathology localisation. 
The reference-based comparison paradigm effectively provides in-context anatomical priors, allowing the model to identify deviations from healthy structure rather than relying on learned disease representations. 
%This explains consistent improvements across diverse VLMs.

\section{Conclusion}

We introduced WALDO, a training-free approach for VLM anomaly localisation grounded in optimal transport theory. Key contributions include: (1) entropy-weighted Sliced Wasserstein reference selection using DINOv2 patch distributions to preserve spatial structure; (2) a theoretical characterisation of the Goldilocks effect explaining why moderate-similarity references optimise bias-variance tradeoff; (3) self-consistency aggregation via weighted NMS for robust bounding box prediction. WALDO achieves 43.5$_{\pm1.6}$\% mAP@30 on NOVA with Qwen2.5-VL-72B (+19.5\% relative improvement, $p<0.01$), with consistent gains across GPT-4o (+68\%), Qwen3-VL-32B (+57\%), and Gemini-2.0-Flash (+110\%). Evaluation on VinDr-CXR also showed promising zero-shot capabilities. Our results show that providing healthy reference images enables VLMs to perform differential diagnosis via in-context anatomical priors, advancing zero-shot anomaly detection towards clinical utility.

\begin{credits}
\subsubsection{\ackname} We acknowledge HPC resources from NHR@FAU (projects b143dc, b180dc), funded by federal and Bavarian state authorities and Gerhard Wellein's and his team's HPC approach. NHR@FAU hardware is partially funded by DFG 440719683. Additional support was received from ERC projects MIA-NORMAL 101083647, CHARMS 101246053, and ORACLE 101326871, DFG 513220538 and 512819079, and the state of Bavaria (HTA and the Bavarian Foundation Model Initiative). We further acknowledge resources provided by the Isambard-AI National AI Research Resource (AIRR), operated by the University of Bristol and funded by DSIT via UKRI and STFC [ST/AIRR/I-A-I/1023]~\cite{mcintoshsmith2024isambardai}. We used coding agents and LLMs from Anthropic, OpenAI, Google, and Mistral AI, for text polishing, coding, experiment orchestration, and cluster monitoring. Most remarkable was the models’ rigorous support with falsification of weaker hypotheses through extensive testing and mathematical scrutiny.

\subsubsection{\discintname}
The authors have no competing interests to declare that are relevant to the content of this article.
\end{credits}

%\section*{Acknowledgments}
%Anonymous for review.

\bibliographystyle{splncs04}
\bibliography{references}

\clearpage

\section*{Supplementary Material}

\begin{figure}[h]
\centering
\includegraphics[width=\textwidth]{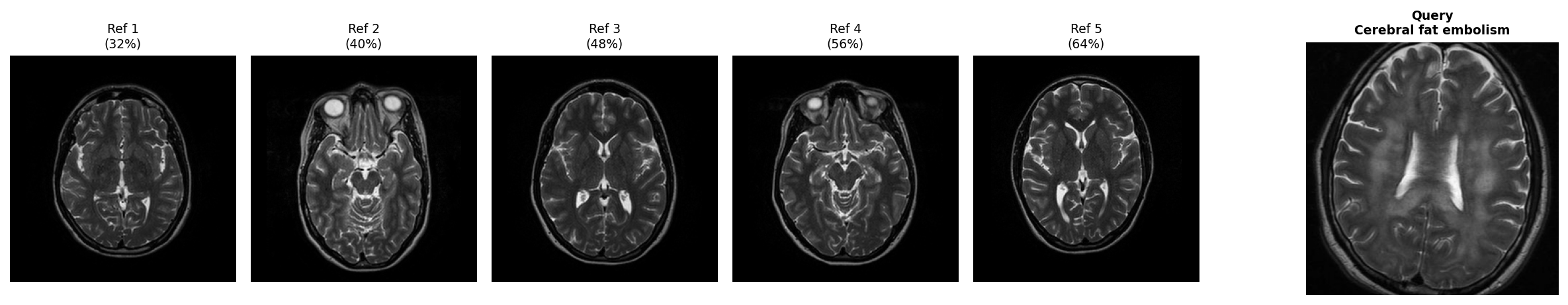}

\vspace{0.3em}
\fbox{
\begin{minipage}{0.95\textwidth}
\scriptsize
\textbf{System:} \texttt{You are a medical imaging expert analysing brain MRI scans.}

\vspace{0.2em}
\textbf{User:} \texttt{I will show you 5 healthy brain MRI reference images, followed by a query image that may contain pathology. Compare the query image to the healthy references. Identify any regions that appear abnormal, \emph{i.e.}, areas with unusual intensity, texture, or structure compared to the healthy examples. Return bounding boxes in JSON format: [\{"box\_2d": [y1, x1, y2, x2], "label": "description"\}]. Coordinates should be normalised to [0, 1000].}

\vspace{0.2em}
\texttt{[5 reference images shown above, then query image]}

\vspace{0.2em}
\hrule
\vspace{0.2em}

\textbf{Stage 1 Response:} \texttt{[\{"box\_2d": [180, 285, 420, 510], "label": "multiple hyperintense foci"\}]}

\textbf{Stage 2 Refinement:} \texttt{Given your prediction, examine the query more carefully and refine if needed.}

\textbf{Stage 2 Response:} \texttt{[\{"box\_2d": [195, 290, 405, 495], "label": "cerebral fat embolism"\}]}
\end{minipage}
}
\caption{WALDO prompt example: Cerebral fat embolism (\textbf{IoU=0.928}). Top: 5 Goldilocks-selected IXI healthy references (axial T2, 32\%--64\% $SW_2^{(w)}$ similarity) and query image. Bottom: full prompt structure with two-stage VLM response. The Goldilocks selection (30-70th percentile similarity) intentionally picks moderately similar references rather than the closest matches - this provides useful contrast for the VLM to identify what's ``different'' without overfitting to specific anatomical structures.}
\label{fig:prompt_v16}
\end{figure}

\begin{figure}[h]
\centering
\includegraphics[width=\textwidth]{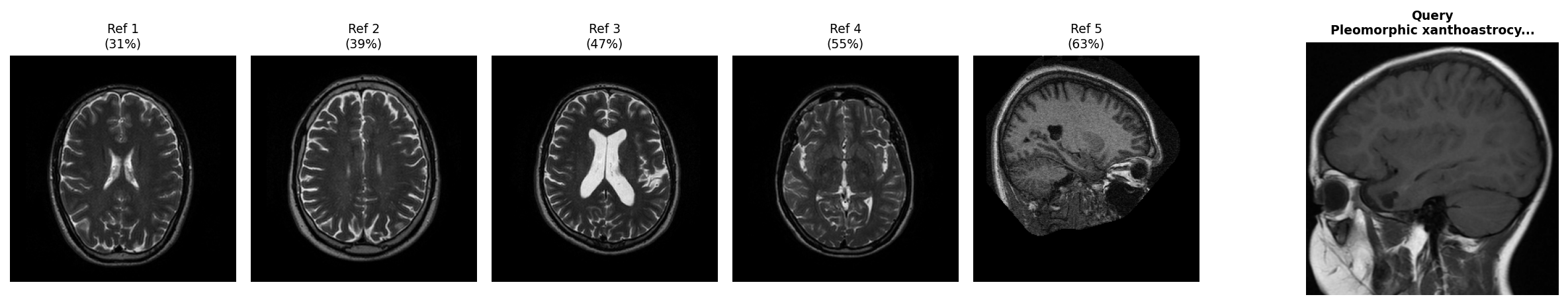}

\vspace{0.3em}
\fbox{
\begin{minipage}{0.95\textwidth}
\scriptsize
\textbf{System:} \texttt{You are a medical imaging expert analysing brain MRI scans.}

\vspace{0.2em}
\textbf{User:} \texttt{I will show you 5 healthy brain MRI reference images, followed by a query image that may contain pathology. Compare the query image to the healthy references. Identify any regions that appear abnormal, \emph{i.e.}, areas with unusual intensity, texture, or structure compared to the healthy examples. Return bounding boxes in JSON format: [\{"box\_2d": [y1, x1, y2, x2], "label": "description"\}]. Coordinates should be normalised to [0, 1000].}

\vspace{0.2em}
\texttt{[5 reference images shown above, then query image]}

\vspace{0.2em}
\hrule
\vspace{0.2em}

\textbf{Stage 1 Response:} \texttt{[\{"box\_2d": [85, 320, 380, 590], "label": "heterogeneous mass with cystic components"\}]}

\textbf{Stage 2 Refinement:} \texttt{Given your prediction, examine the query more carefully and refine if needed.}

\textbf{Stage 2 Response:} \texttt{[\{"box\_2d": [95, 335, 365, 575], "label": "pleomorphic xanthoastrocytoma"\}]}
\end{minipage}
}
\caption{WALDO prompt example: Pleomorphic xanthoastrocytoma with dural reaction. Top: 5 Goldilocks-selected healthy references from mixed IXI+OASIS pool (31\%--63\% $SW_2^{(w)}$ similarity) and query image. Bottom: full prompt structure with two-stage VLM response. The VLM provides both localisation coordinates and a zero-shot disease label based on visual features.}
\label{fig:prompt_v19}
\end{figure}

\begin{figure}[h]
\centering
\includegraphics[width=\textwidth]{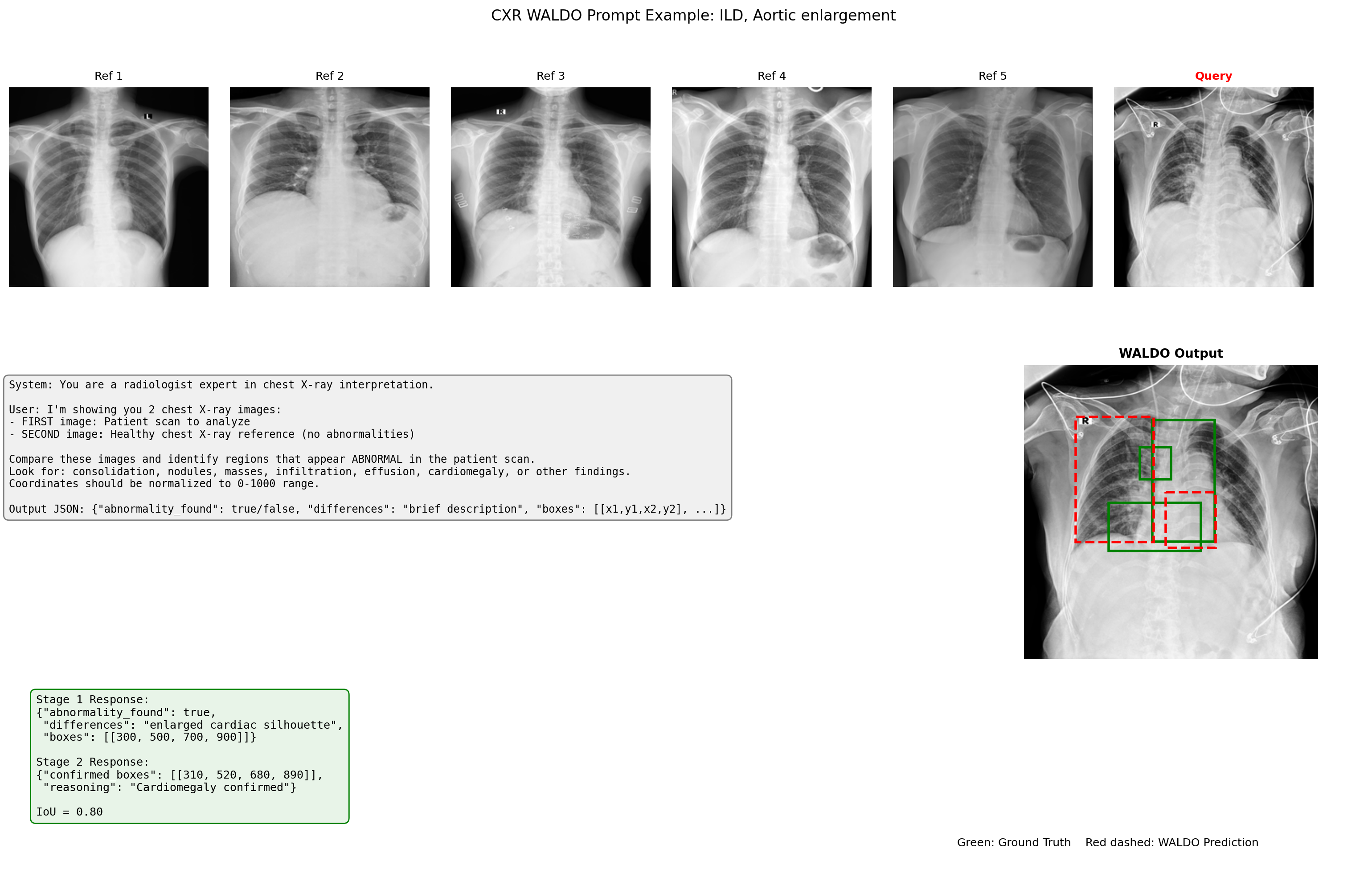}
\caption{CXR WALDO prompt example: Cardiomegaly (IoU=0.80). Top row: 5 Goldilocks-selected healthy references from ``No Finding'' cases, followed by the clean query image (patient CXR). Middle: Differential prompt structure. Bottom right: WALDO output with predictions (green: ground truth, red dashed: WALDO predictions). The differential prompting compares the patient scan to healthy references to identify abnormal regions.}
\label{fig:cxr_prompt}
\end{figure}

Figures~\ref{fig:prompt_v16}, \ref{fig:prompt_v19}, and \ref{fig:cxr_prompt} show examples for WALDO prompts. 

% === MIXED IXI + OASIS REFERENCE EXAMPLES ===

\end{document}